\documentclass[twocolumn]{article}
\usepackage[utf8]{inputenc}
\usepackage[]{graphicx}
\usepackage[backend=bibtex]{biblatex}
\usepackage{float}
\usepackage{afterpage}
\usepackage{algorithm}
\usepackage[noend]{algpseudocode}
\usepackage{amsmath}
\makeatletter
\def\BState{\State\hskip-\ALG@thistlm}
\makeatother

\title{Group Visual Sentiment Analysis}
\author{Zeshan Hussain, Tariq Patanam and Hardie Cate}
\date{June 6, 2016}

\begin{document}

\maketitle

\begin{abstract}
     \textit{In this paper, we introduce a framework for classifying images according to high-level sentiment. We subdivide the task into three primary problems: emotion classification on faces, human pose estimation, and 3D estimation and clustering of groups of people. We introduce novel algorithms for matching body parts to a common individual and clustering people in images based on physical location and orientation. Our results outperform several baseline approaches.}
\end{abstract}

\section{Introduction}

A key problem in computer vision is sentiment analysis. Uses include identifying sentiment for social network enhancement and emotion understanding by robotic systems. Identifying individual people and faces has many uses but has been explored extensively.  Group sentiment analysis, on the other hand, has been much less researched. Given an image of a group of people, we may wish to describe the overall sentiment of the scene. For example, if we have an image of students in a classroom, we may wish to determine the level of human interaction, happiness of the students, their degree of focus, and other soft scene characteristics that together describe the overall sentiment. To tackle this problem, we propose a multi-label classification system that outputs labels for each of our scene characteristics. To perform this classification, we localize dense features from faces and poses of people as well as spatial relations of people in the image.

\section{Related Work} 

\subsection{Summary of Previous Work}
Individual sentiment analysis has been a long studied problem. There exists a hosts of methods to extract emotion features. Notable among them is the Half-Octave Multi-Scale Gaussian derivative technique to quickly extract . Employing this method, Jain and Crowley were able to accurately detect over 90\% smiles accurately \cite{jain2013smile}

Discovering groups in images itself is a significant problem. Detecting people around occlusions and viewpoint changes, and then grouping according to orientations and the poses of the people for multiple groups of people is a highly complex problem that has only recently been solved. Specifically, Choi et al. describe a model that learns an ensemble of discriminative interaction patterns to encode the relationships between people in 3D \cite{choi2014discovering}. Their model learns a minimization potential function to efficiently map out groups of people in images as well as their activity and pose (i.e. standing, facing each other). Prior work has focused on the activity of a single person or pairs of people \cite{laptev2005space} \cite{dollar2005behavior} \cite{niebles2008unsupervised} \cite{liu2009recognizing} as well as pose estimation of groups, albeit operating under the assumption that there is only one group in the image \cite{eichner2010we}.

There has also been some work done analyzing the sentiment of groups in images. Dhall et al., for example, measure happiness intensity levels in groups of people by utilizing a model that takes advantage of the global and local attributes of the image \cite{dhall2015automatic}. Borth et al. take a more general approach to sentiment analysis by training several concept classifiers on millions of images taken from Flickr. A prediction for an arbitrary image is an adjective-noun pair that defines the scene as well as its sentiment \cite{borth2013large}.

Finally, work has been done analyzing crowds of people by taking a more top-down approach. Using behavior priors, Rodriguez et al. track the motion of crowds. Other methods include taking a bottom-up approach, where collective motion patterns are learned and used to constrain the motion of individuals within crowds in a certain test scene \cite{rodriguez2011data}.

\subsection{Improvements on Previous Work}
We contribute a novel multi-label group sentiment classification system that is built by using several components from previous papers, including emotion and pose classification as well as group structure and orientation prediction. Our method of combining these features to classify group sentiment is a new approach to sentiment analysis. Additionally, we propose a simple, novel clustering approach to predict group structure and orientation. Specifically, we use a heuristic for 3D estimation of the people in the image and then use a variant of k-means to cluster them.

\section{Technical Approach}

\subsection{Summary}
For each input image, $I$, we perform the following feature extractions: 

\begin{enumerate}
    \item Emotion Extraction 
    \item Poselet Estimation 
    \item Group Structure and Orientation Estimation 
\end{enumerate}

Suppose that each feature extraction results in the feature vectors, $f_1$, $f_2$, $f_3$, respectively. Our extraction is done in such a way that for a fixed $i=1,2,3$, the number of entries in $f_i$ is the same for all images. This allows us to simply concatenate features from each of the three approaches for use in an SVM classification.

\subsection{Technical Background and Datasets}

Our primary dataset consists of 597 images, each corresponding to one of six scenes which are "bus stop," "cafeteria," "classroom," "conference," "library," and "park." It is the same dataset used by \cite{choi2014discovering}. We define four scene characteristics, namely level of human interaction, physical activity, happiness, and focus, that describe each image. In order to provide a ground truth labeling for image characteristics, we go through all of our images and manually annotate them on a scale of 1 to 4, with 1 as the least and 4 as the most. For instance, a classroom scene with students working at their desks would most likely correspond to a focus score of 4, whereas an image of people staring off into space while waiting for a bus would likely have a score of 1.

\subsection{Technical Solution}

To get a better sense for the problem, we begin with several (naive) baseline approaches that we improve upon with more advanced techniques. First, we create a color histogram for each image that we use as input features for an SVM. Each pixel consists of values for 3 color channels and we divide the possible values for channel into 8 bins, so each pixel votes for one of $8\times8\times8=512$ bins according to the values in its color channels. We then use this histogram of 512 values as input for an SVM. This model ends up overfitting on our small dataset but performs slightly better than random on a validation set.

As a second baseline, we incorporate features from bounding boxes that define the locations of people in the images. These bounding boxes come with the dataset which was used in the paper by Choi et al. discussed above. We use the coordinates of the bounding boxes directly as features for input to an SVM. We limit the number of bounding boxes per image to 15 (since the number of bounding boxes in most images were below this threshold) by randomly selecting boxes if there are more present in the image and padding our flattened vector of coordinates with zeros if there are fewer boxes in the image. Similar to the above approach, this technique only slightly outperforms a random selection process, with more detailed results discussed below.

\subsection{Emotion Classification}

Our emotion classification pipeline involves several stages.

\subsubsection{Face Detection}

First, we detect faces by leveraging the Poselet detector \cite{bourdev2009poselets} and the Haar Cascade classifier built into OpenCV \cite{opencv_library}. 
For a given image, we get all face-related poselets (effectively bounding boxes within the image) from the Poselet detector and then run the Haar Cascade classifier to detect faces within these bounding boxes.
We remove duplicate faces by checking for overlaps among the face bounding boxes.

Unfortunately, the dataset we use contains many people who are far away from the camera and so their faces are often small and therefore difficult to detect with traditional face detection methods. To address this issue, we take advantage of the fact that we have bounding boxes from people from \cite{choi2014discovering} and that we can identify torsos of people using poselets. This additional information makes it possible to adjust the settings of the Haar classifier to be lenient and then prune erroneous faces using a novel matching algorithm that we describe below.

Suppose we have an image $I$. Suppose further that we have sets of bounding boxes in the image for people, faces, and torsos which we denote $P$, $F$, and $T$, respectively. First, we loop over $F$ and assign to each face $f\in F$ a person $p\in P$ such that $f$ is contained in $p$ and $p$ is the bounding box of the ones remaining that minimizes the distance between the centers of the top edges of the bounding boxes. This distance is a useful heuristic because faces should be centered and near the top within the bounding box to which they belong.

Next, we loop over $P$ from smallest to largest and assign to each person $p\in P$ a torso $t\in T$ such that $t$ is contained in $p$ and $t$ is the largest such torso from among the remaining unassigned torsos. We proceed through $P$ from smallest to largest because smaller bounding boxes are less likely to have a large number of torso contained in them, so these bounding boxes provide tighter restraints. We choose the largest torso contained in $P$ because we find that there are often a large number of small erroneous torsos while larger torso (that also fit within a bounding box) are more likely to correspond to actual human torsos.

\begin{figure}[ht]
    \centering
    \includegraphics[width=\linewidth]{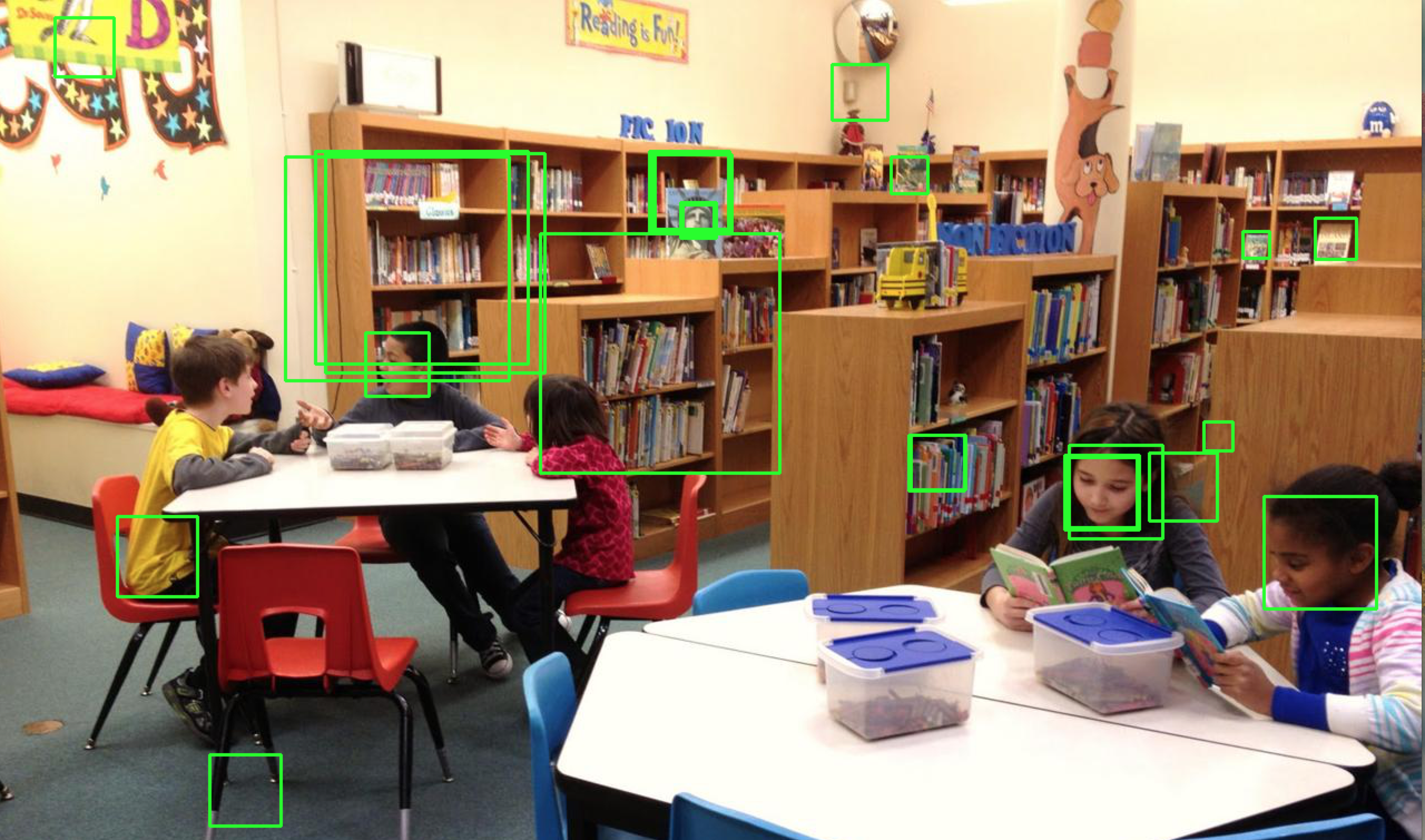}
    \includegraphics[width=\linewidth]{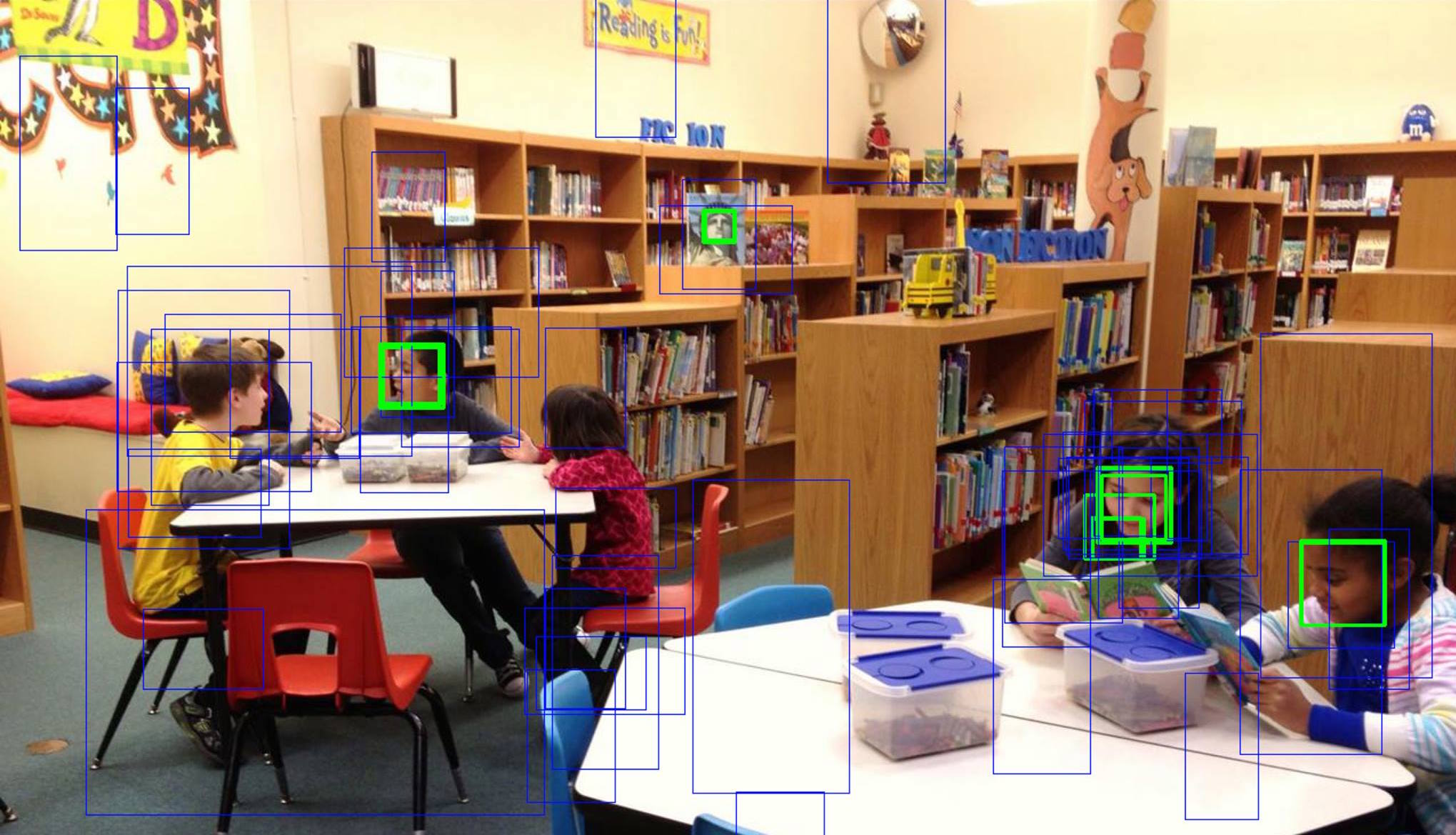}
    \caption{Face extraction when done only with a Haar Cascade face detector (top) vs. first looking for head poselets and then detecting faces within them}
    \label{fig:bbmatching}
\end{figure}

At the end of the algorithm, each bounding box corresponding to a person is assigned at most one face and one torso. 
Any unassigned bounding box for a person, face, or torso is discarded. This allows us to concentrate on the most important people in the image (i.e., people nearest to and likely facing the camera).
Torsos are potentially useful later on for pose estimation using silhouettes. Faces are useful both for emotion classification and for 3D estimation.

\subsubsection{Emotion Classification Method}

Once the faces are detected, we classify them with emotion labels.
In order to do so, we generate features for each face using a Multi-scale Gaussian pyramid modeled after the approach found in \cite{jain2013smile}.
Given a face image as input, we convert the image to grayscale and iteratively perform Gaussian convolutions on the image at varying scales. 
At each iteration, we compute first- and second-order image gradients at each pixel, then aggregate the means and standard deviations of these gradients in 4x4 windows across the entire image.
Each 4x4 window at a given scale produces 10 features, 5 from the means of the five different gradient types and 5 from the standard deviations.
At each scale, we aggregate features three times using the half-octave method, and we build our pyramid to a depth of three scales. At scale $i$, we perform the following convolution pipeline to an input image. 

\begin{align*}
\mathrm{Given:\ } I_{0}(x,y,i) \\
I_{1}(x,y,i) &= I_{0}(x,y,i)*g(x,y;\sigma) \\
I_{2}(x,y,i) &= I_{1}(x,y,i)*g(x,y;\sqrt{2}\sigma) \\
I_{0}(x,y,i+1) &= I_{2}(x,y,i)
\end{align*}
We use the aggregate of $I_{0}$, $I_{1}$, $I_{2}$ at each stage as our features.

Though this technique is originally used strictly for detecting smiles, we extend the technique to classify other emotions since the same features should be relevant.
Using the feature vectors produced from the Multi-scale Gaussian pyramid, we train an SVM on these features to classify various emotions. 
We train on a dataset that contains only faces and is annotated with emotions including happy, sad, surprised, fear, etc.
A more complete discussion can be found in the Experiments section, but ultimately we use a binary happy/neutral classification when training our SVM.

Once we have a prediction for each face, we divide the image into 16 evenly spaced windows. Each window acts as 2-class histogram to produce a count of the number of smiles and neutral faces for each portion of the image. We then use this  32-element vector as our feature vector for our emotion pipeline. This binning process standardizes the length of the feature vector for each image so that we can run an SVM on these features for each of our final sentiment classes.


\subsection{Pose Estimation}

As mentioned earlier, the Poselet detector provides a wealth of information relating to people in images. There are 150 types of poselet descriptors, and each one acts as a detector for different body parts (e.g., left leg, right arm, head, shoulders, etc.). 

Given an input image, we get all poselets as bounding boxes along with an i.d. and score for each poselet. The i.d. identifies the type of poselet and the score indicates the accuracy with which the image within the bounding box matches that particular poselet type. We filter out all poselets below a certain score threshold. Empirically, we find that a score of 0.9 provides a reasonable threshold. After filtering out these erroneous poselets, we produce a 150-class feature vector for the image by binning poselets according to their i.d. This gives us a count of the number of occurrences of each poselet type.

As discussed in the Experiments section, these feature vectors provide a reasonable estimation of the different human poses throughout the image. Poselets not only identify body parts, but they also give indications of the shapes of these parts (e.g., elbow bent vs. elbow straight). These feature vectors from poselets ultimately prove to have significant predictive power.

\subsection{Group Structure and Orientation Estimation}

We now describe our approach for predicting orientation and group structures in images. First, we compare two methods for prediction of orientation. We then propose a 3D estimation algorithm to estimate the 3D coordinates of people. Then, using both orientations and 3D coordinates, we run a variant of the k-means algorithm to cluster people into groups. Below, we detail our two methods for orientation prediction. 

In the first method, we use grab cut to get the silhouettes of the people in each image. To estimate relative foreground and background of each person, we utilize the matched torso and face per bounding box. Specifically, we mark a stripe of pixels in the face of the person and a rectangular area of pixels in the torso of a person to be the foreground and all the pixels outside of the bounding box to be the background. We then extract SIFT features on the resultant silhouettes and train an SVM using these features to predict orientation.  

Our second method is slightly simpler. Instead of extracting SIFT features from silhouettes, we extract HOG features from the bounding box. The benefit of this second method is that we obtain a denser feature representation of the image without losing the "global" nature of the image. Because HOG looks at the image more globally, it does better with occlusions. For our orientation prediction, we move forward with this second method due to its simplicity, representational efficiency, and time constraints.

To estimate the depth of each person, we utilize the following relationship between depth and size of the face, $d = \frac{k}{f}$, where $d$ is depth, $f$ is the size of the face in the image, and $k$ is a constant. Intuitively, as the face size decreases, the depth of the face increases. 

Finally, with each person mapped to a 3D coordinate, we can run a variant of $k$-means constrained by the orientations of people to find clusters of groups in the image. Our variation of $k$-means uses an orientation coefficient which linearly weights the traditional Euclidean distance between a point and its corresponding cluster centroid.
For each person with unit orientation vector $\theta$, we compute the vector $\theta - \phi$, where $\phi$ is the unit vector along the direction between the person and the cluster center. 
The orientation coefficient $c$ is computed as a linear function of $\theta - \phi$, with $c=0.5$ when $\theta = \phi$ and $c=1.5$ when $\theta = - \phi$.
This orientation heuristic favors clusters in which more people in a cluster are facing the cluster center, mimicking situations in which people are facing each other during a conversation.

With this modified distance function in place, we run $k$-means with varying values of $k$. We then choose the value of $k$ that minimizes the sum of modified distances between points and their cluster centers while placing a restriction on the number of singleton clusters (clusters with only one person). We place this restriction on singleton clusters because we want to avoid the situation in which all centers are coincident with people since we do not know the value of $k$ beforehand. Note that we do not use mean shift because we also do not know the appropriate window size beforehand, and mean shift generally does not perform as well on small, sparse datasets. We define our potential function below which we want to maximize to achieve the optimal $k$.

\begin{align*}
    \sum_{i}^{n} \vec{o} \cdot \vec{c} - k * \sum_{i}^{n} d
\end{align*}

where

\begin{align*}
    k &= \mathrm{constant\ factor} \\
    o &= \mathrm{orientation\ vector} \\
    c &= \mathrm{direction\ vector\ to\ cluster\ center} \\
    d &= \mathrm{distance\ to\ cluster\ center}
\end{align*}

This allows us to avoid over-invidualizing the groupings while only including persons within the a reasonable distance.

It is important to note that due to time constraints, we were unable to fully implement the 3D estimation and orientation aspects of our pipeline.
We did, however, implement most of the pipeline and were able to produce a very accurate classifier of person orientations.
We believe the approach outlined above would identify groups of people adequately enough to provide additional predictive power on the overall problem of sentiment classification.

\begin{figure*}[h]
  \centering
  \begin{minipage}[t]{0.24\textwidth}
    \includegraphics[width=\textwidth]{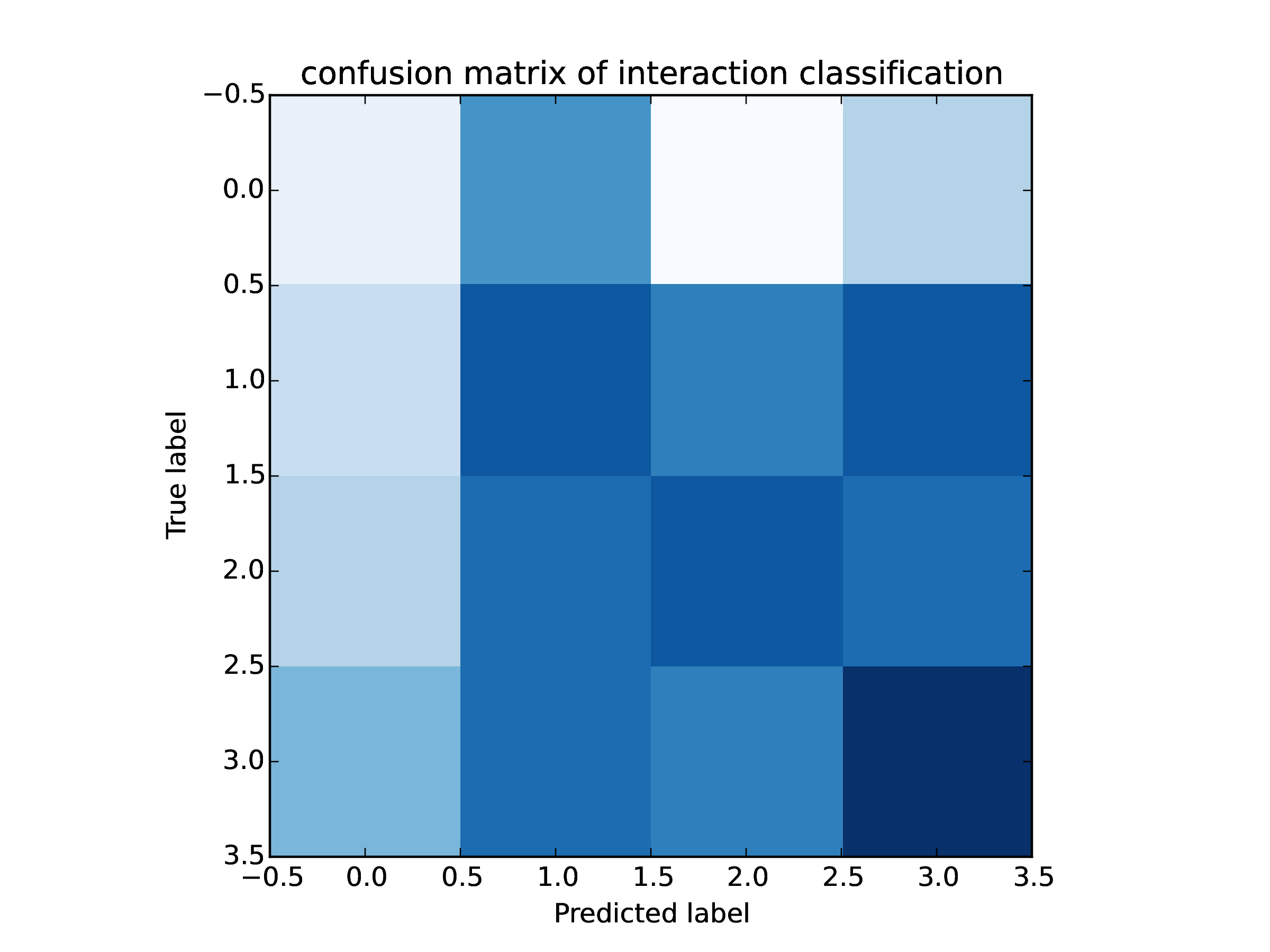}
    \caption{Final confusion matrix of interaction prediction}
    \label{fig:interaction_confusion_final}
  \end{minipage}
  \hfill
  \begin{minipage}[t]{0.24\textwidth}
    \includegraphics[width=\textwidth]{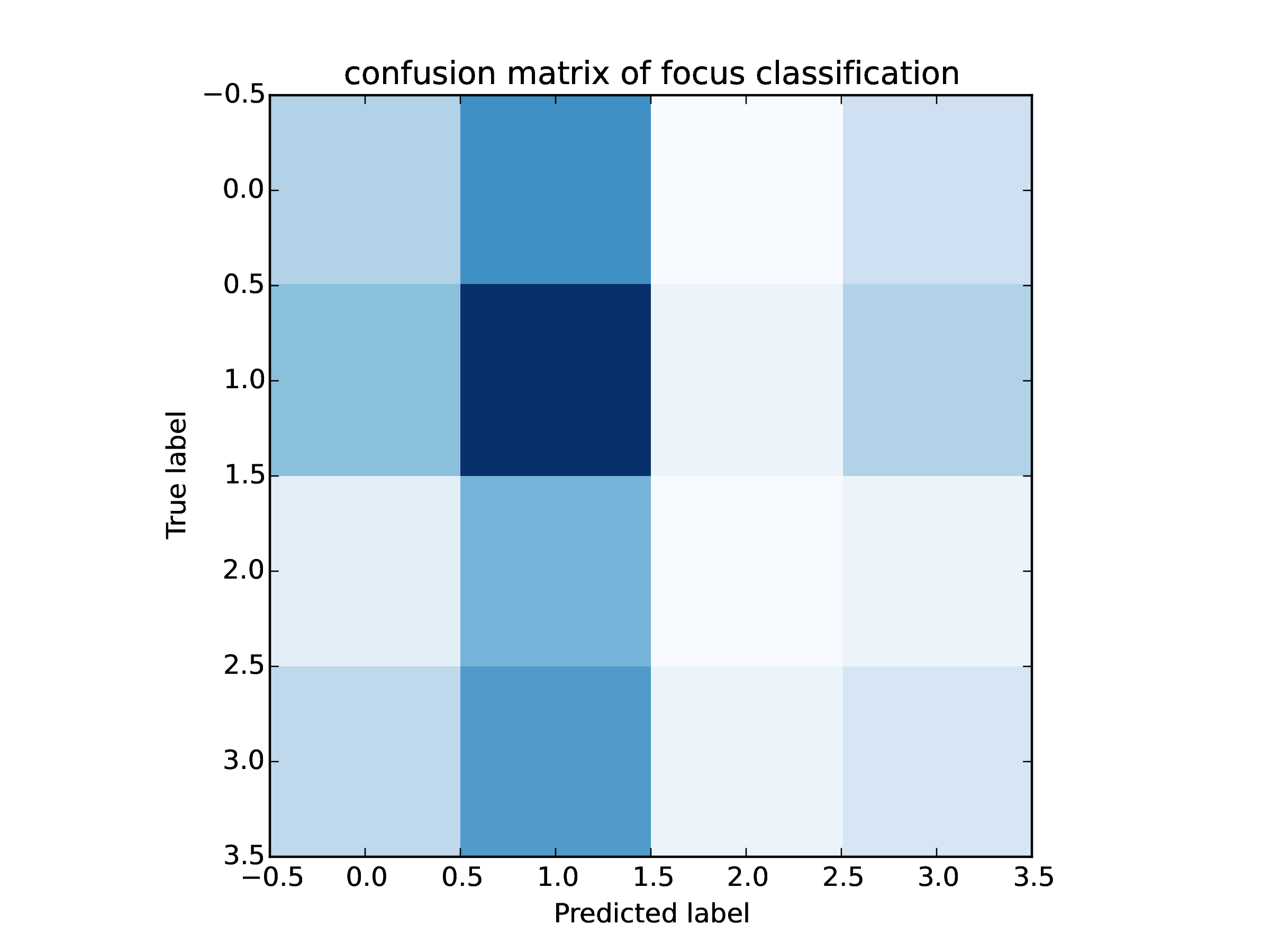}
    \caption{Final confusion matrix of focus prediction}
    \label{fig:focus_confusion_final}
  \end{minipage}
  \hfill
  \begin{minipage}[t]{0.24\textwidth}
    \includegraphics[width=\textwidth]{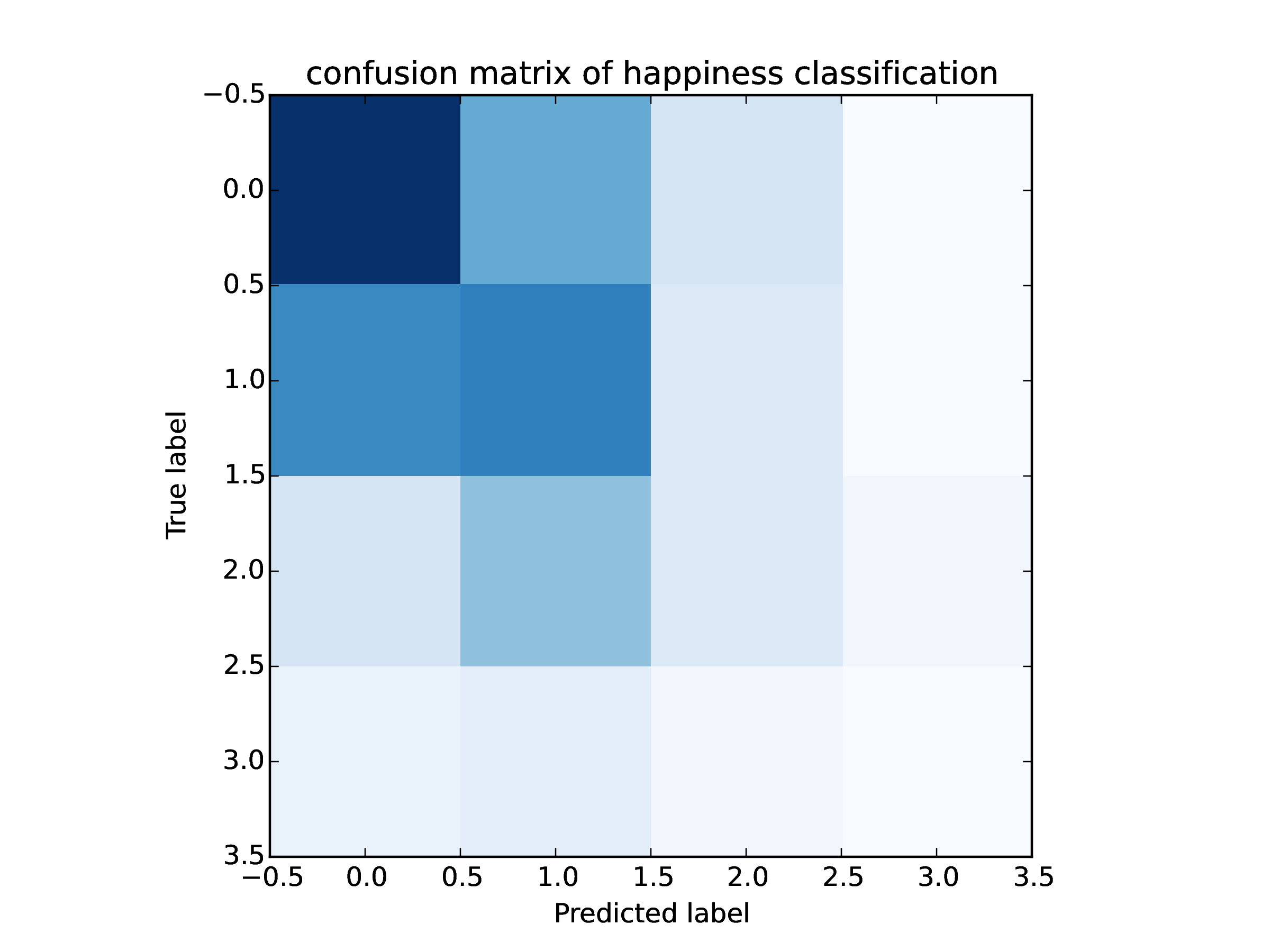}
    \caption{Final confusion matrix of happiness prediction}
    \label{fig:happiness_confusion_final}
  \end{minipage}
  \hfill
  \begin{minipage}[t]{0.24\textwidth}
    \includegraphics[width=\textwidth]{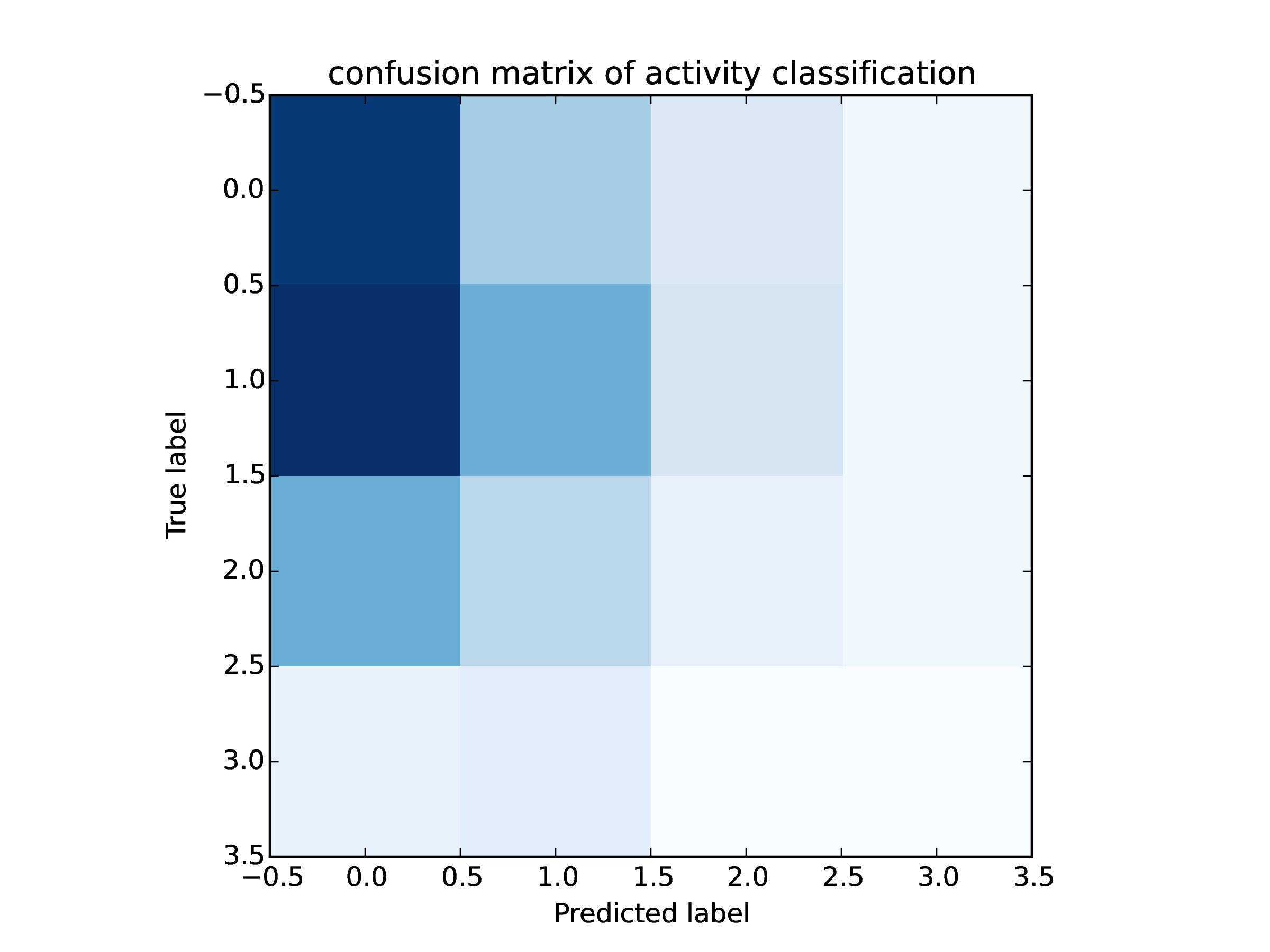}
    \caption{Final confusion matrix of activity prediction}
    \label{fig:activity_confusion_final}
  \end{minipage}
\end{figure*}

\section{Experiments}

As mentioned, getting a grasp of the group sentiment requires several individualized parts to come together. Therefore, as to bring together these individualized parts, we conducted a series of experiments. 

\subsection{Baselines}
We start by establishing two baselines. Our first baseline is built by extracting color histograms for each images then modeling an SVM using those feature vectors. We achieve 28.3\% or about 3\% better than random. Our second baseline is built by building a feature vector from the bounding boxes. The idea behind this was to form a crude estimation of the groupings. It performed relatively poorly, achieving an average accuracy of 30\%, only 5\% better than randomly binning into one of four intensities for a sentiment class. 

\subsection{Smile Detection}
A key component of group sentiment is determining individual sentiment. We begin by testing our implementation of the Half-Octave Multi-Scale Guassian Pyramid algorithm\cite{jain2013smile} on the GENKI 4K Dataset\cite{GENKI-4K}. We successfully achieve an accuracy of 83.5\% when classifying in binary fashion for smile or neutral, on par with current methods for smile detection. 

\begin{table}[h]
    \centering
    \begin{tabular}{|l|c|c|}
    \hline
        & Crowley & Us\\ \hline
         Training error &  N/A & 0.0 \\
         Testing error & 0.082 & 0.165\\ \hline
    \end{tabular}
    \caption{Error comparisons with Crowley paper}
    \label{tab:smiledetection}
\end{table}

Admittedly, our accuracy is slightly below that of Jain \& Crowley as we do not fine-tune the parameters of the soft-margin SVM as meticulously. Yet as we will see this individualized emotion extraction algorithm plays a key roll in improving overall group sentiment analysis prediction.
\subsection{Sad and Happy Detection with the Kaggle Dataset}

Using the same Gaussian Pyramid algorithm, we further test our ability to extract features and detect emotions on the Kaggle Facial Expressions Recognition Challenge dataset. The detection system worked less robustly on this set achieving 71.5\% accuracy in classifying between happiness and neutrality and 56\% when classifying between sadness and neutrality.

There are two important observations to note here. First, the images in the GENKI set are more refined, featuring almost always only faces that are larger than those from the Kaggle set. The Kaggle set meanwhile often has pictures with hands covering parts of the face. Occlusions, then, are particularly challenging to the Multi-Scale Gaussian method. Second, the Gaussian Pyramid algorithm seems particularly well-adjusted to the intricacies of a smile. While not performing as well when identifying "happy" verus "neutral" in the Kaggle set as it did on the GENKI set, the algorithm did better within the Kaggle set on the binary classification of "happy" versus the binary classification of "sad." This could be because the sadness expressed in the Kaggle dataset was usually visually less expressive than the happiness, and therefore less distinct from the neutral state  


\subsection{Orientation Classification}

The next stage in overall group sentiment analysis is incorporating the groupings between the individual people. In order to do that, we built a relatively robust SVM model using HOG feature extraction trained on the TUD Multiview Pedestrian Dataset, successfully achieving 90.3\% accuracy when classifying orientations into one of eight cardinal directions. Such a high accuracy rate is likely due to the fact that gradient orientations can easily define the overall orientation of people, and that gradient orientations are the basis of HOG extracted features. 

\begin{figure}[ht]
    \centering
    \includegraphics[width=0.3\linewidth]{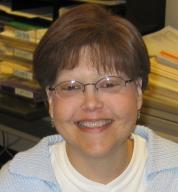}
    \includegraphics[width=0.3\linewidth]{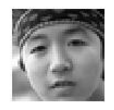}
    \includegraphics[width=0.3\linewidth]{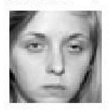}
    \caption{A comparison of a GENKI happy, a neutral Kaggle, and a sad Kaggle image. Note there is little visual difference between the neutral and sad Kaggle images}
    \label{fig:genki_kaggle_comparison}
\end{figure}

\subsection{Group Sentiment Anlaysis}

We can then featurize our (detailed in the our approach) distinct features into an overall scene feature vector. Their performance is detailed below. 

\subsubsection{Individual Emotion Extraction}

When using an overall facial emotion feature vector by, we achieve strong results for happiness and activity. This is expected as our Multi-Scale Gaussian algorithm proves strongest in being able to detect smiles which correlate heavily with the overall happiness of a scene as well as, in the case of the Group Discovery Dataset, with activity. In contrast, overall group "focus" is probably more correlated with posture and "interaction" more correlated with the number and size of the groupings. 

\subsubsection{Poselets}

When using only the most identifying poses visible in the scene, weighing them by their score from the poselet detector, we achieve better results only for "activity." While we expect better results for focus as well, this could very well be an indication that "focus" is not easily measurable from one frame of a scene. 

\subsubsection{Individual Emotion Extraction and Poselets}

Our final experiment is our true goal: testing our cohesive method for classifying overall scene sentiment analysis. For each sentiment we bin into four intensities and achieve accuracies as follows:

\begin{table*}[t]
    \centering
    \begin{tabular}{|p{2cm}||p{2cm}|p{2cm}|p{2cm}|p{2cm}|}
     \hline
     \multicolumn{5}{|c|}{Combined Feature Extraction Results} \\
     \hline
     Metric& Happiness& Activity& Interaction&Focus \\
     \hline
     Training Error    &0.306 &0.285 &0.310 &0.266\\
     \hline
     Testing Error &0.558  &0.617 &0.642 &0.667\\
     \hline
    \end{tabular}
    \caption{Errors for the SVM trained by extracting emotions and poselets}
    \label{table:baseline2_accuracies}
\end{table*}


\subsubsection{A Binary Classification of Intensities}

We achieve 45\% accuracy in classifying the happiness of scenes into their right intensity bucket. However, classifying other sentiment values were less accurate. To a great degree, this could be because of the fickleness of human labeling. It is often hard, even for a human, to distinguish between a focus intensity of 3 and a focus intensity of 4. As such, we ran one final experiment, labeling all intensity values of 3 or 4 as 1 and all intensity values of 1 or 2 as 0. The following are the results for this modified binary classifier. 

\begin{table*}[t]
    \centering
    \begin{tabular}{|p{2cm}||p{2cm}|p{2cm}|p{2cm}|p{2cm}|}
     \hline
     \multicolumn{5}{|c|}{Combined Feature Extraction Results under a Binary Classifier} \\
     \hline
     Metric& Happiness& Activity& Interaction&Focus \\
     \hline
     Training Error &0.189 &0.163 &0.237 &0.277\\
     \hline
     Testing Error &0.233  &0.242 &0.4 &0.4167\\
     \hline
    \end{tabular}
    \caption{Errors for the SVM trained by extracting emotions and poselets but using a binary classifier}
    \label{table:baseline2_accuracies}
\end{table*}

Taking happiness as an example, as it is our most accurate result, we achieve a 53.4\% improvement over random selection while with an intensity classifier we achieve an 80.4\% improvement over random classification. Thus there is not a strong improvement in using a binary classifier over the four degree intensity classifier which means the weaknesses more likely, not in the gradients of labeling but in the previously mentioned techniques of feature extraction itself. 

\subsection{Final Notes}

Unfortunately we did not have enough time to use our robust orientation detection model to test group discovery. An ordering of our thoughts, however, are detailed in the technical approach. In order to achieve even better results in the future, we would likely need to incorporate more complex scene information such as objects in the background and scene type.

\section{Conclusion}
In closing, we summarize a couple of important results. For our feature extraction results, we have high performance detecting emotion (namely, smiles) as well as on predicting orientation. Using these features, we perform reasonably on happiness and activity, We see a strong correlation between our extracted emotions with happiness level as well as between poselets and activity level in the image, which is an intuitive result. In our further work, we hope to complete our group estimation pipeline and combine these features into our final feature vectors. Adding the group features will most likely increase the accuracy of our system in predicting interaction and focus. 

In this paper, we contribute a multi-label classification system that performs significantly above random chance with our feature extraction methods. Our approach effectively utilizes previous feature extraction methods as well as novel methods. 

Finally here is a link to our Github repo: \texttt{https://github.com/zeshanmh/VisualSentiment}

\section{References}
\nocite{*}
\printbibliography[heading=none]

\end{document}